\documentclass{llncs}

\usepackage{amsfonts}
\usepackage{graphicx}
\usepackage{multirow}

\begin{document}
\title{Suggestive Annotation: A Deep Active Learning Framework for Biomedical Image Segmentation}
\author{Lin Yang$^1$, Yizhe Zhang$^1$, Jianxu Chen$^1$, Siyuan Zhang$^2$, Danny Z. Chen$^1$}
\institute{$^1$ Department of Computer Science and Engineering,\\ University of Notre Dame, Notre Dame, IN 46556, USA\\
$^2$ Department of Biological Sciences, Harper Cancer Research Institute,\\ University of Notre Dame, Notre Dame, IN 46556, USA}
	
\maketitle

\begin{abstract}
Image segmentation is a fundamental problem in biomedical image analysis. Recent advances in deep learning have achieved promising results on many biomedical image segmentation benchmarks. However, due to large variations in biomedical images (different modalities, image settings, objects, noise, etc), to utilize deep learning on a new application, it usually needs a new set of training data. This can incur a great deal of annotation effort and cost, because only biomedical experts can annotate effectively, and often there are too many instances in images (e.g., cells) to annotate.
In this paper, we aim to address the following question: With limited effort (e.g., time) for annotation, what instances should be annotated in order to attain the best performance?  We present a deep active learning framework that combines fully convolutional network (FCN) and active learning to significantly reduce annotation effort by making judicious suggestions on the most effective annotation areas. We utilize uncertainty and similarity information provided by FCN and formulate a generalized version of the maximum set cover problem to determine the most representative and uncertain areas for annotation. Extensive experiments using the 2015 MICCAI Gland Challenge dataset and a lymph node ultrasound image segmentation dataset show that, using annotation suggestions by our method, state-of-the-art segmentation performance can be achieved by using only 50\% of training data.


\end{abstract}

\section{Introduction}

Image segmentation is a fundamental task in biomedical image analysis. 
Recent advances in deep learning \cite{chen2016deep,chen2016dcan,ronneberger2015u,xu2016gland2,xu2016gland1} have achieved promising results on many biomedical image segmentation benchmarks \cite{10.3389/fnana.2015.00142,sirinukunwattana2017gland}. Due to its accuracy and generality, deep learning has become a main choice for image segmentation.
But, despite its huge success in biomedical applications, deep learning based segmentation still faces a critical obstacle: the difficulty in acquiring sufficient training data due to high annotation efforts and costs.
Comparing to applications in natural scene images, it is much harder to acquire training data in biomedical applications for two main reasons. (1) Only trained biomedical experts can annotate data, which makes crowd leveraging quite difficult. (2) Biomedical images often contain much more object instances than natural scene images, which can incur extensive manual efforts of annotation.
For example, public datasets in biomedical areas have significantly fewer spatial annotated images (85 for MICCAI Gland Challenge \cite{sirinukunwattana2017gland}; 30 for ISBI EM Challenge \cite{10.3389/fnana.2015.00142}). 

To alleviate the common burden of manual annotation, an array of weakly supervised segmentation algorithms \cite{hong2015decoupled} has been proposed. However, they did not address well the question that which data samples should be selected for annotation for high quality performance.
Active learning \cite{settles2010active}, which allows the learning model to choose training data, provided a way to answer this need. As shown in \cite{dutt2016active}, using active learning, state-of-the-art level performance can be achieved using significantly less training data in natural scene image segmentation. But, this method is based on the pre-trained region proposal model and pre-trained image descriptor network, which cannot be easily acquired in biomedical image settings due to large variations in biomedical applications.

In this paper, we present a new framework that combines fully convolutional network (FCN) \cite{long2015fully} and active learning \cite{settles2010active} to reduce annotation effort by making judicious suggestions on the most effective annotation areas. To address the issues in \cite{dutt2016active}, we exploit FCN to obtain domain specific image descriptor and directly generate segmentation without using region proposals.
\begin{figure}[t]
	\centering
	\includegraphics[width=12cm]{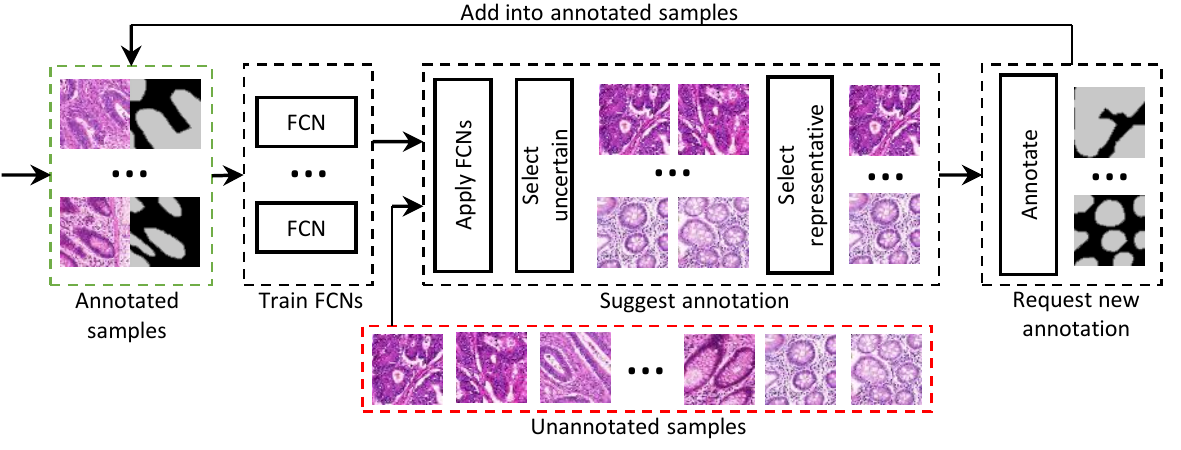}
	\caption[ ]{Illustrating our overall deep active learning framework.}
	\label{fig1}
\end{figure}
Fig.~\ref{fig1} outlines the main ideas and steps of our deep active learning framework. Starting with very little training data, we iteratively train a set of FCNs. At the end of each stage, we extract useful information (such as uncertainty estimation and similarity estimation) from these FCNs to decide what will be the next batch of images to annotate. After acquiring the new annotation data, the next stage is started using all available annotated images.
Although the above process seems straightforward, we need to overcome several challenges in order to integrate FCNs into this deep active learning framework, as discussed below. 

\textbf{Challenges from the perspective of FCNs.} (1) The FCNs need to be fast to train, so that the time interval between two annotation stages is acceptable. (2) They need to be of good generality, in order to produce reasonable results when little training data is available.
To make the model fast to train, we utilize the ideas of batch normalization \cite{ioffe2015batch} and residual networks \cite{he2016deep}. Then, we use bottleneck design \cite{he2016deep} to significantly reduce the number of parameters (for better generality) while maintaining a similar number of feature channels as in \cite{chen2016dcan}.

\textbf{Challenges from the perspective of active learning.} It needs to exploit well the information provided by the FCNs when determining the next batch of training data. For this, we first demonstrate how to estimate uncertainty of the FCNs based on the idea of bootstrapping and how to estimate similarity between images by using the final layer of the encoding part of the FCNs. Based on such information, we formulate a generalized version of the maximum set cover problem \cite{feige1998threshold,hochbaum1996approximating} for suggesting the next batch of training data.

Experiments using the 2015 MICCAI Gland Challenge dataset \cite{sirinukunwattana2017gland} and a lymph node ultrasound image segmentation dataset \cite{zhang2016coarse} show that (1) annotation suggestions by our framework are more effective than common methods such as random query and uncertainty query, and (2) our framework can achieve state-of-the-art segmentation performance by using only 50\% of training data. 


\section{Method}
Our proposed method consists of three major components: (1) a new FCN, which shows state-of-the-art performance on the two datasets used in our experiments; (2) uncertainty estimation and similarity estimation of the FCNs; (3) an annotation suggestion algorithm for selecting the most effective training data.

\begin{figure}[t]
	\centering
	\includegraphics[width=11.8cm]{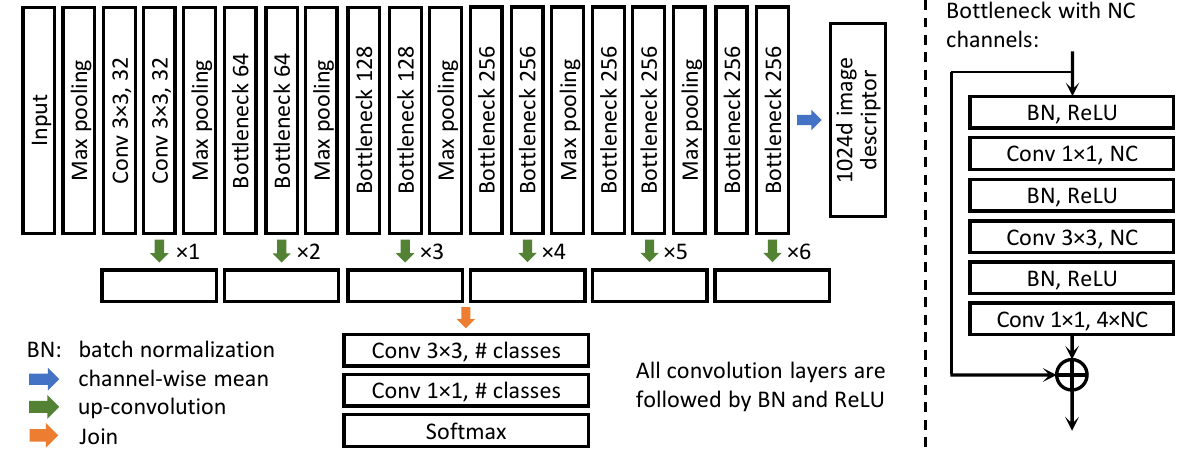}
	\caption[ ]{Illustrating the detailed structure of our FCN components.}
	\label{fig2}
\end{figure}

\subsection{A new fully convolutional network \label{FCN}}
Based on recent advances of deep neural network structures such as batch normalization \cite{ioffe2015batch} and residual networks \cite{he2016deep}, we carefully design a new FCN that has better generality and is faster to train.

Fig.~\ref{fig2} shows the detailed structure of our new FCN.
Its encoding part largely follows the structure of DCAN \cite{chen2016dcan}.
As shown in both residual networks \cite{he2016deep} and batch normalization \cite{ioffe2015batch}, a model with these modifications can achieve the same accuracy with significantly fewer training steps comparing to its original version.
This is essential when combining FCNs and active learning, since training FCNs usually takes several hours before reaching a reasonable performance.
Thus, we change the original convolution layers into residual modules with batch normalization.
Note that, at the start of active learning, since only few training samples are available, having too many free parameters can make the model hard to train. Hence, we utilize the bottleneck design \cite{he2016deep} to reduce the number of parameters while maintaining a similar number of feature channels at the end of each residual module.
In the decoding part of the network, we modify the structure in \cite{chen2016deep} to gradually enlarge the size of the feature maps to ensure a smooth result.
Finally, a $3\times3$ convolution layer and a $1\times1$ convolution layer are applied to combine the feature maps from different scales together.
As the experiments show, our new FCNs can achieve state-of-the-art performance when all training data is used while still able to produce reasonable results when very little training data is available.

\begin{figure}[t]
	\centering
	\includegraphics[width=11cm]{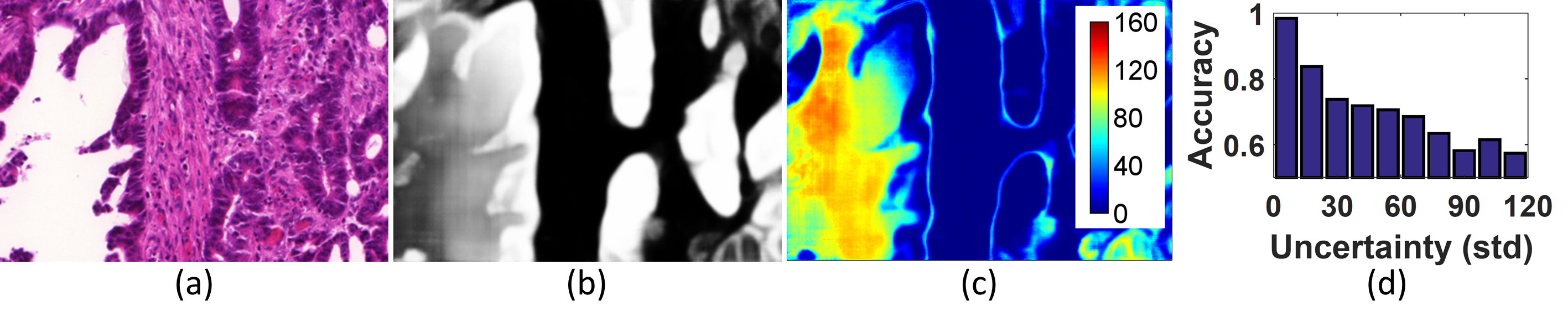}
	\caption[ ]{(a) An original image; (b) the probability map produced by our FCNs for (a); (c) uncertainty estimation of the result; (d) relation between uncertainty estimation and pixel accuracy on the testing data. This shows that the test accuracy is highly correlated with our uncertainty estimation.
	}
	\label{fig3}
\end{figure}

\subsection{Uncertainty estimation and similarity estimation \label{estimation}}
A straightforward strategy to find the most ``valuable" annotation areas is to use uncertainty sampling, with the active learner querying the most uncertain areas for annotation.
However, since deep learning models tend to be uncertain for similar types of instances, simply using uncertainty sampling will result in duplicated selections of annotation areas.
To avoid this issue, our method aims to select not only uncertain but also highly representative samples (samples that are similar to lots of other training samples).
To achieve this goal, we need to estimate the uncertainty of the results and measure the similarity between images. In this section, we illustrate how to extract such information from FCNs.

\begin{figure}[t]
	\centering
	\includegraphics[width=11cm]{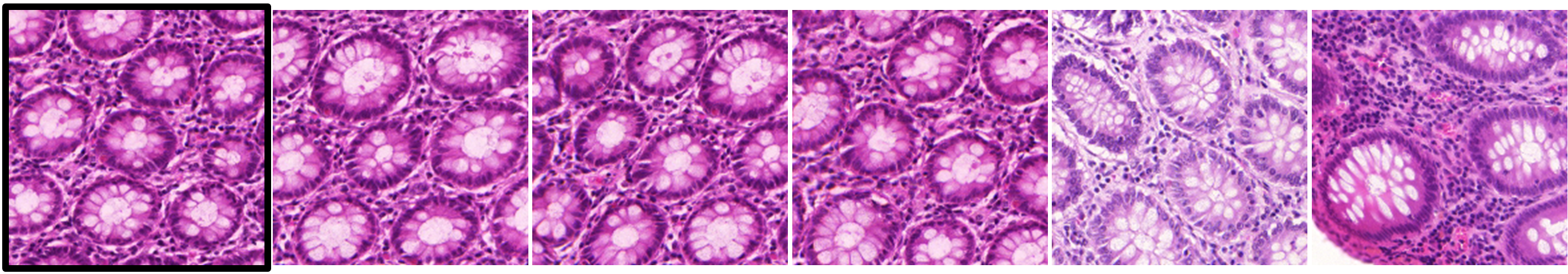}
	\caption[ ]{Illustrating similarity estimation: The 5 images on the right have the highest similarity scores with respect to the leftmost images among all training images in \cite{sirinukunwattana2017gland}. 
	}
	\label{fig4}
\end{figure}

Bootstrapping \cite{efron1994introduction} is a standard way for evaluating the uncertainty of learning models. Its basic idea is to train a set of models while restricting each of them to use a subset of the training data (generated by sampling with replacement) and calculate the variance (disagreement) among these models. We follow this procedure to calculate the uncertainty of FCNs. Although the inner variance inside each FCN can lead to overestimation of the variance, in practice, it can still provide a good estimation of the uncertainty. As shown in Fig.~\ref{fig3}(d), the estimated uncertainty for each pixel has a strong correlation with the testing errors. Thus, selecting uncertain training samples can help FCNs to correct potential errors. Finally, the overall uncertainty of each training sample is computed as the mean uncertainty of its pixels.

CNN based image descriptor has helped produce good results in natural scene images.
The encoding part of FCN is naturally an CNN, and for an input image $I_i$, the output of the last convolution layer in the encoding part can be viewed as high level features $I^f_i$ of $I_i$.
Next, to eliminate shifting and rotation variances of the image, we calculate the channel-wise mean of $I^f_i$ to generate condensed features $I^c_i$ as the domain-specific image descriptor.
This approach has two advantages.
(1) There is no need to train another separate image descriptor network.
(2) Because the FCNs are trying to compute the segmentation of the objects, $I^c_i$ contains rich and accurate shape information. 
Finally, we define the similarity estimation between two images $I_i$ and $I_j$ as: $sim(I_i,I_j) = \mbox{\textit{cosine\_similarity}}(I^c_i,I^c_j)$.
Fig.~\ref{fig4} shows an example of the similarity estimation.

\subsection{Annotation suggestion\label{Annotation_suggestion}}
To maximize the effectiveness of the annotation data, the annotated areas are desired to be typical or representative in terms of the following two properties.
(1) Uncertainty: The annotated areas need to be difficult to segment for the network.
(2) Representativeness: The annotated areas need to bear useful characteristics or features for as many unannotated images as possible.
In this section, we show how to suggest a set of areas for annotation that very well satisfy these two properties, based on similarity estimation and uncertainty estimation.

In each annotation suggestion stage, among all unannotated images, $\mathcal{S}_{u}$, we aim to select a subset of $k$ images, $\mathcal{S}_{a} \subseteq \mathcal{S}_{u}$, that is both highly uncertain and representative.
Since uncertainty is a more important criterion, in step 1, images with the top $K$ ($K>k$) uncertainty scores are extracted and form a candidate set $\mathcal{S}_{c}$.
In step 2, we find $\mathcal{S}_{a} \subseteq \mathcal{S}_{c}$ that has the largest representativeness.

To formalize the representativeness of $\mathcal{S}_{a}$ for $\mathcal{S}_{u}$, we first define the representativeness of $\mathcal{S}_{a}$ for an image $I_{x} \in \mathcal{S}_{u}$ as: 
$f(\mathcal{S}_{a},I_{x}) = \max_{I_{i}\in\mathcal{S}_{a}} sim(I_{i},I_{x})$, where $sim(\cdot,\cdot)$ is the similarity estimation between $I_{i}$ and $I_{x}$. 
Intuitively, $I_{x}$ is represented by its most similar image in $\mathcal{S}_{a}$, 
measured by the similarity $sim(\cdot,\cdot)$.
Then, we define the representativeness of $\mathcal{S}_{a}$ for $\mathcal{S}_{u}$ as: 
$F(\mathcal{S}_{a},\mathcal{S}_{u}) = \sum_{I_{j} \in \mathcal{S}_{u}} f(\mathcal{S}_{a},I_{j})$, which reflects how well $\mathcal{S}_{a}$ represents all the images in $\mathcal{S}_{u}$.
By finding $\mathcal{S}_{a} \subseteq \mathcal{S}_{c}$ that maximizes $F(\mathcal{S}_{a},\mathcal{S}_{u})$, we promote $\mathcal{S}_{a}$ by
(1) selecting $k$ ``hub'' images that are similar to many unannotated images and
(2) covering diverse cases (since adding annotation to the same case does not significantly increase $F(\mathcal{S}_{a},\mathcal{S}_{u})$).

Finding $\mathcal{S}_{a} \subseteq \mathcal{S}_{c}$ with $k$ images that maximizes $F(\mathcal{S}_{a},\mathcal{S}_{u})$ can be formulated as a generalized version of the maximum set cover problem \cite{feige1998threshold}, as follows. We first show when $sim(\cdot,\cdot)\in \{0,1\}$, the problem is an instance of the maximum set cover problem. For each image $I_i \in \mathcal{S}_{c}$, $I_i$ covers a subset $\mathcal{S}_{I_i}\subseteq \mathcal{S}_{u}$, where $I_y\in \mathcal{S}_{I_i}$ if and only if $sim(I_i,I_y)=1$. Further, since $sim(\cdot,\cdot)\in \{0,1\}$, for any $I_x\in \mathcal{S}_{u}$, $f(\mathcal{S}_{a},I_{x})$ is either 1 (covered) or 0 (not covered) and $F(\mathcal{S}_{a},\mathcal{S}_{u})$ (the sum of $f(\mathcal{S}_{a},I_{x})$'s) is the total number of the covered images (elements) in $\mathcal{S}_{u}$ by $\mathcal{S}_{a}$. Thus, finding a $k$-images subset $\mathcal{S}_{a} \subseteq \mathcal{S}_{c}$  maximizing $F(\mathcal{S}_{a},\mathcal{S}_{u})$ becomes finding a family $\mathcal{F}$ of $k$ subsets from $\{S_{I_i} \ | \ I_i \in \mathcal{S}_{c}\}$ such that $\cup_{S_j \in \mathcal{F}} S_j$ covers the largest number of elements (images) in $\mathcal{S}_{u}$ (max $k$-cover \cite{feige1998threshold}). The maximum set cover problem is NP-hard and its best possible polynomial time approximation algorithm is a simple greedy method \cite{feige1998threshold} (iteratively choosing $S_i$ to cover the largest number of uncovered elements). Since our problem is a generalization of this problem (with $sim(\cdot,\cdot)\in [0,1]$, instead of $sim(\cdot,\cdot)\in \{0,1\}$), our problem is clearly NP-hard, and we adopt the same greedy method. Initially, $\mathcal{S}_{a} = \emptyset$ and $F(\mathcal{S}_{a},\mathcal{S}_{u}) = 0$. Then, we iteratively add $I_{i} \in \mathcal{S}_{c}$ that maximizes $F(\mathcal{S}_{a}\cup I_{i},\mathcal{S}_{u})$ over $\mathcal{S}_{a}$, until $\mathcal{S}_{a}$ contains $k$ images. Note that, due to the max operation in $f(\cdot,\cdot)$, adding an (almost) duplicated $I_{i}$ does not increase $F(\mathcal{S}_{a},\mathcal{S}_{u})$ by much. It is easy to show that this algorithm achieves an approximation ratio of $1-\frac{1}{e}$ \cite{hochbaum1996approximating}.


\begin{table}[t]
    \setlength{\tabcolsep}{7pt}
    \renewcommand{\arraystretch}{1.2}
	\centering
	\caption{Comparison with full training data for gland segmentation.}
	\begin{tabular}{| c | c | c | c | c | c | c |}
	  \hline
	  \multirow{2}{*}{Method} & \multicolumn{2}{c|}{F1 score} & \multicolumn{2}{c|}{ObjectDice} & \multicolumn{2}{c|}{ObjectHausdorff} \\
	  \cline{2-7}
	  & Part A & Part B & Part A & Part B & Part A & Part B \\
	  \hline
	  Our method & \textbf{0.921} & \textbf{0.855} & 0.904 & \textbf{0.858} & 44.736 & \textbf{96.976} \\
	  \hline
	  Multichannel \cite{xu2016gland1} & 0.893 & 0.843 & \textbf{0.908} & 0.833 & \textbf{44.129} & 116.821 \\
	  \hline
	  Multichannel \cite{xu2016gland2} & 0.858 & 0.771 & 0.888 & 0.815 & 54.202 & 129.930 \\
	  \hline
      CUMedVision \cite{chen2016dcan} & 0.912 & 0.716 & 0.897 & 0.781 & 45.418 & 160.347 \\
      \hline
	\end{tabular}
	\label{gland}
\end{table}

\section{Experiments and Results}

\begin{figure}[t]
	\centering
	\includegraphics[clip, trim=2.5cm 0.1cm 2.5cm 0.1cm, width=11cm]{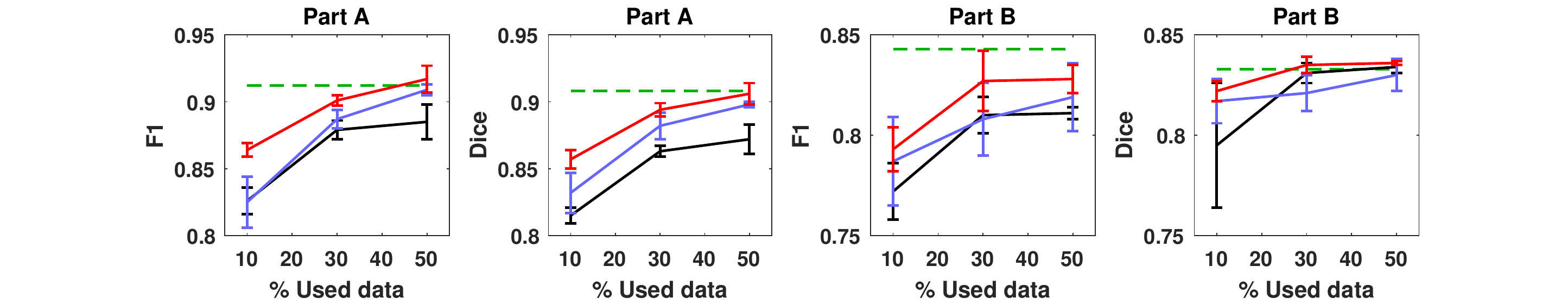}
	\caption[ ]{Comparison using limited training data for gland segmentation: The black curves are for the results of random query, the blue curves are for the results of uncertainty query, the red curves are for the results by our annotation suggestion, and the dashed green lines are for the current state-of-the-art results using full training data.
	} 
	\label{fig5}
\end{figure}

\begin{table}[t]
    \setlength{\tabcolsep}{1.5pt}
    \renewcommand{\arraystretch}{1.2}
	\centering
	\caption{Results for lymph node ultrasound image segmentation.}
	\begin{tabular}{| c | c | c || c | c | c |}
	\hline
	Method & Mean IU & F1 score & Method & Mean IU & F1 score\\
	\hline
	U-Net \cite{ronneberger2015u} & 0.798 & 0.775 & Uncertainty 50\% & 0.858 & 0.849\\
	\hline
	CUMedNet \cite{chen2016deep} & 0.816 & 0.798 & Our method 50\% & 0.875 & 0.871\\
	\hline
	CFS-FCN \cite{zhang2016coarse} & 0.851 & 0.843 & Our method full & \textbf{0.879} & \textbf{0.874}\\
	\hline
	\end{tabular}
	\label{lymph}
\end{table}

To thoroughly evaluate our method on different scenarios, we apply it to the 2015 MICCAI Gland Challenge dataset and a lymph
node ultrasound image segmentation dataset \cite{zhang2016coarse}. The MICCAI data have 85 training images and 80 testing images (60 in Part A; 20 in Part B).
The lymph node data have 37 training images and 37 testing images. In our experiments, we use $k=8$, $K=16$, $2000$ training iterations, and $4$ FCNs. The waiting time between two annotation suggestion stages is 10 minutes on a workstation with 4 NVIDIA Telsa P100 GPU.
We use 5\% of training data as validation set to select the best model.

\textbf{Gland segmentation.} We first evaluate our FCN module using full training data.
%
As Table~\ref{gland} shows, on the MICCAI dataset, our FCN module achieves considerable improvement on 4 columns ($\sim2\%$ better), while has very similar performance on the other two ($\sim0.5\%$ worse). Then, we evaluate the effectiveness of our annotation suggestion method, as follows. To simulate the annotation suggestion process, we reveal training annotation only when the framework suggests it. The annotation cost is calculated as the number of revealed pixels. Once the annotation cost reaches a given budget, we stop providing more training data. In our experiment, we set this budget as 10\%, 30\%, and 50\% of the overall labeled pixels. We compare our method with (1) random query: randomly requesting annotation before reaching the budget, and (2) uncertainty query: selecting annotation areas based only on uncertainty estimation ($K=k$). Fig.~\ref{fig5} summarizes the results. It shows that our annotation suggestion method is consistently better than random query and uncertainty query, and our framework can achieve state-of-the-art performance using only 50\% of the training data.

\textbf{Lymph node segmentation.} Table~\ref{lymph} summarizes the results on lymph node segmentation. ``Our method full'' entry shows the results of our FCN using all training data. ``Our method 50\%'' and ``Uncertainty 50\%'' entries show the comparison between uncertainty query and our annotation suggestion method under the 50\% budget. It shows that our framework achieves better performance in all cases. By using 50\% of the training data, our framework attains better segmentation performance than the state-of-the-art method \cite{zhang2016coarse}.

\section{Conclusions}
In this paper, we presented a new deep active learning framework for biomedical image segmentation by combining FCNs and active learning. Our new method provides two main contributions: (1) A new FCN model that attains state-of-the-art segmentation performance; (2) an annotation suggestion approach that can direct manual annotation efforts to the most effective annotation areas.
\\
\\
\textbf{Acknowledgment.} This research was supported in part by NSF Grants CCF-1217906, CNS-1629914, CCF-1617735, CCF-1640081, NIH Grant 5R01CA194697-03, and the Nanoelectronics Research Corporation, a wholly-owned subsidiary of the Semiconductor Research Corporation, through Extremely Energy Efficient Collective Electronics, an SRC-NRI Nanoelectronics Research Initiative under Research Task ID 2698.005.

\bibliographystyle{splncs03}
\bibliography{reference}
\end{document}